%% file: coling_mcqa.tex
\pdfoutput=1

\documentclass[11pt]{article}

\usepackage[final]{coling}

\usepackage{times}
\usepackage{latexsym}

\usepackage[T1]{fontenc}
\usepackage[utf8]{inputenc}
\usepackage{microtype}

\usepackage{inconsolata}

\usepackage{graphicx}
\usepackage{pifont}
\usepackage{multirow}
\usepackage{multicol}
\usepackage{subfigure}
\usepackage{booktabs}
\usepackage{amsmath}
\usepackage{amssymb}

\title{LLMs May Perform MCQA by Selecting the Least Incorrect Option}

\author{Haochun Wang, Sendong Zhao$\thanks{ \ \ Corresponding author}$, Zewen Qiang, Nuwa Xi, Bing Qin and Ting Liu
 \\
Research Center for Social Computing and Information Retrieval, \\Harbin Institute of Technology, China \\
\texttt{\{hcwang,sdzhao\}@ir.hit.edu.cn} \\
}

\begin{document}
\maketitle
\begin{abstract}
   \input{Text/0.abstract} 
\end{abstract}

\input{Text/1.introduction}
\input{Text/2.related_work}
\input{Text/3.invariability_does_not}
\input{Text/4.least_incorrect}

\input{Text/6.MCQA_plus}

\input{Text/7.conclusion}

\clearpage
\newpage

\input{Text/8.limitation_n_ethics}
\input{Text/8.z_acknowledgement}

\bibliography{bib}

\appendix

\end{document}

%% file: Text/0.abstract.tex
In the field of NLP, Large Language Models (LLMs) have markedly enhanced performance across a variety of tasks.
However, the comprehensive evaluation of LLMs remains an inevitable challenge for the community.
Recently, the adoption of Multiple Choice Question Answering (MCQA) as a benchmark for assessing LLMs has gained considerable traction.
However, concerns regarding the robustness of this evaluative method persist.
Building upon previous discussions on the issue of \textit{variability}, we reveal an additional dimension of concern:
LLMs may perform MCQA by selecting the least incorrect option rather than distinctly correct.
This observation suggests that LLMs might regard multiple options as correct, which could undermine the reliability of MCQA as a metric for evaluating LLMs.
To address this challenge, we introduce an enhanced dataset augmentation method for MCQA, termed MCQA+,
to provide a more accurate reflection of the model performance,
thereby highlighting the necessity for more sophisticated evaluation mechanisms in the assessment of LLM capabilities.

%% file: Text/1.introduction.tex
\section{Introduction}
The emergence of Large Language Models (LLMs), such as GPT-3~\cite{gpt3}, LLaMA~\cite{touvron2023llama},
and ChatGPT~\cite{chatgpt}, represents a paradigm shift in the field of Natural Language Processing (NLP).
These models have exhibited exceptional proficiency in mimicking human-like textual outputs,
establishing their significance across various applications.
However, the challenge of effectively evaluating LLMs persists~\cite{chang2023survey}.
This difficulty arises from the intricate nature of natural language.
Conventional evaluation metrics for generative tasks often fall short in accurately assessing the performance of LLMs,
since most LLMs can generate text contextually rich and coherent~\cite{thoppilan2022lamda},
complicating the assessment of the outputs through merely quantitative evaluation based on text matching such as BLEU~\cite{papineni-etal-2002-bleu} and ROUGE~\cite{lin-2004-rouge}.

\begin{figure}[ht] 
\centering 
\includegraphics[width=0.99\columnwidth]{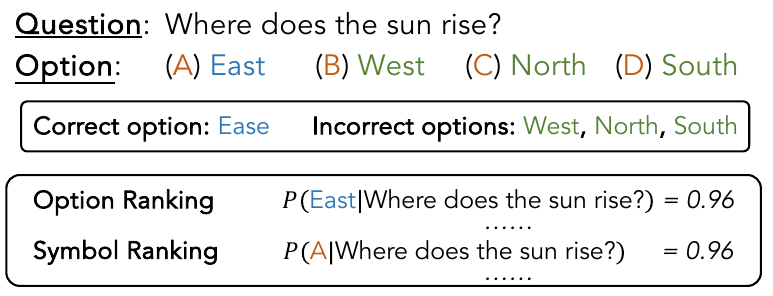}
\caption{An MCQA example and ranking strategies.} 
\label{figure:intro_mcqa} 
\end{figure}

Multiple-Choice Question Answering (MCQA) is a fundamental format for various tasks in NLP,
such as commonsense reasoning~\cite{csqa,siqa,swag}, reading comprehension~\cite{race,cqa} and cloze-style tasks~\cite{hs,sc}.
Each MCQA instance comprises a question paired with several answer options,
requiring models to identify the correct response as depicted in Figure~\ref{figure:intro_mcqa}.
As a non-subjective metric, MCQA serves as a prominent automatic evaluation method with accuracy as an evaluation metric for numerous LLMs to test for the commonsense knowledge or knowledge for specific domain~\cite{leo_gao_2021_5371629,touvron2023llama,openai2023gpt4}.

\begin{figure}[t]
\centering
\includegraphics[width=0.99\columnwidth]{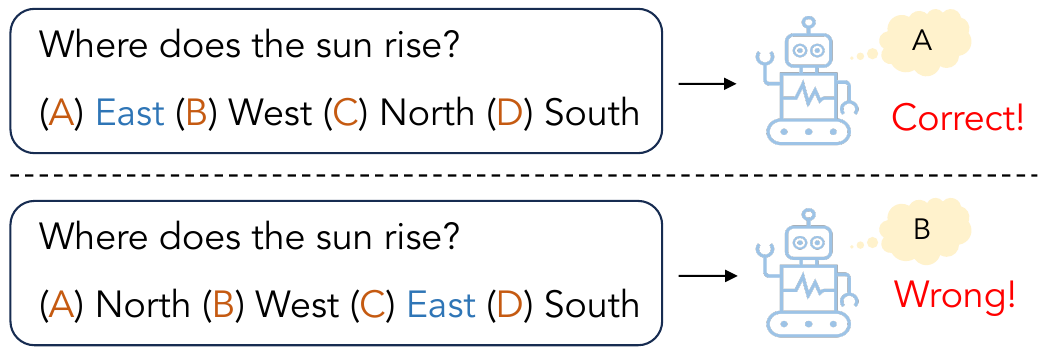}
\caption{A case for variability issue of LLMs.}
\label{figure:case_varia}
\end{figure}
Despite the advanced performance of LLMs on the accuracy of MCQA-format benchmarks like MMLU~\cite{mmlu},
previous studies have discussed a key challenge that persists in evaluating LLMs is maintaining \textit{invariability} in responses when confronted with different orders of answer choices for a same question~\cite{robinson2022leveraging,wang2023large,zheng2023large}, which underscores an issue that the accuracy of MCQA-format tasks may not reflect the authentic capability of LLMs as an example in Figure \ref{figure:case_varia}. However, the above phenomenon may not be the \textit{only} issue in the evaluation of LLMs with MCQA-format questions.

To eliminate the potential impact of variability in model responses, we begin by filtering a dataset, denoted as $\mathcal{D}$, to extract a subset $\mathcal{D}^\blacklozenge$, which contains instances where the LLMs can consistently predict the correct answer across all permutations of the answer options, thereby demonstrating invariability. Following this, we conduct a comprehensive experimental analysis using various configurations derived from the original MCQs in $\mathcal{D}^\blacklozenge$. Our experimental results indicate that while LLMs often select the \textit{most} correct answer, they may also regard other options as correct to some extent. Consequently, evaluating LLM performance solely based on MCQA can produce ambiguous results. This newly identified issue prompts a reconsideration of the suitability of MCQA as a reliable metric for LLM evaluation and offers a possible explanation for the observed differences in LLM performance on generative versus discriminative tasks \cite{west2023generative}.

To address this issue, which is inherently difficult to resolve, we propose an augmentation of the MCQA dataset, termed MCQA+, which introduces variations of the original MCQs and is designed to more accurately reflect LLM capabilities. Empirical findings demonstrate that LLM performance on the MCQA+ dataset is significantly lower than on the original MCQA dataset, indicating that MCQA+ can serve as a more effective benchmark for developing robust and adaptable NLP models. This augmentation may ultimately contribute to narrowing the gap between machine learning models and human-like understanding and reasoning in NLP tasks.

In summary, our contributions are as follows:
\begin{itemize}
\item We identify a novel issue with MCQA-based evaluation of LLMs, beyond the variability in answer options, where LLMs may approach MCQA by selecting the option that is ``least incorrect''.
\item This issue implies that while LLMs consistently select the correct answer for specific MCQs, they may also incorrectly identify certain other options as correct under different circumstances. 
\item We introduce a dataset augmentation method, expanding the original MCQA into MCQA+, which more accurately reveals LLM capacities and performance. 
\end{itemize}

%% file: Text/2.related_work.tex
\begin{figure*}[tbp] 
\centering 
\includegraphics[width=0.99\textwidth]{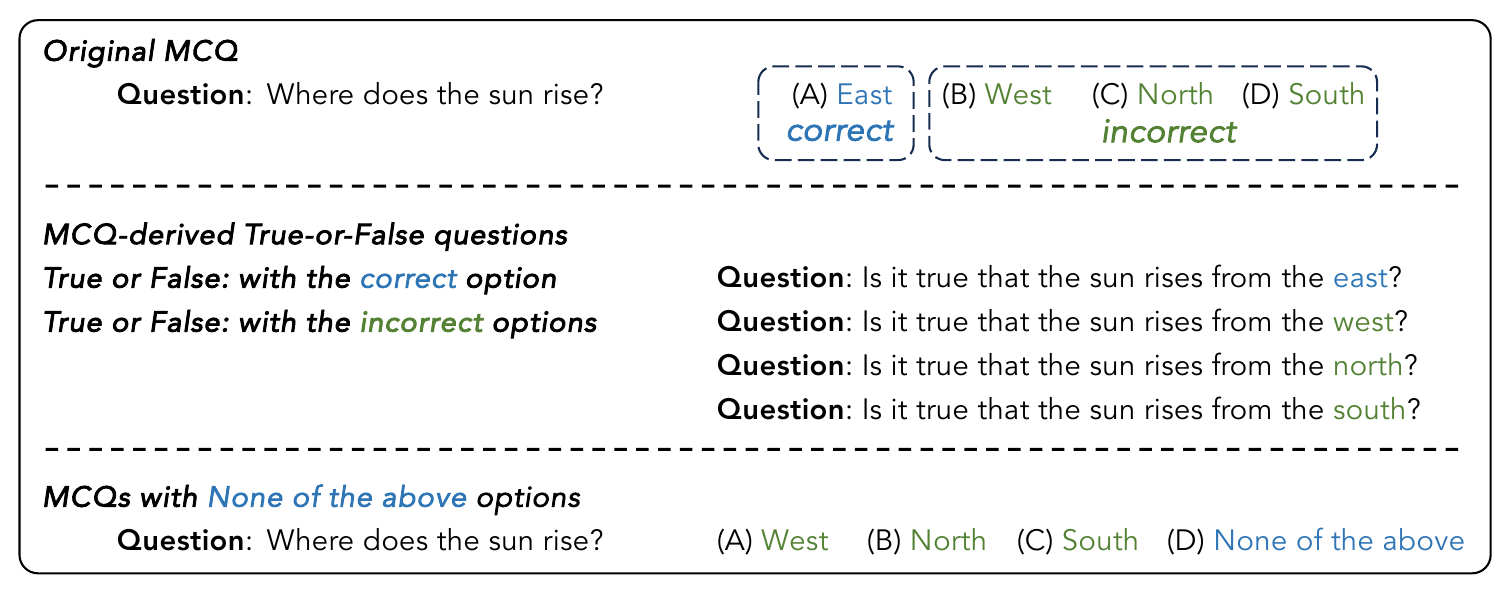}
\caption{A case of an original MCQ, True-or-false questions derived from the MCQ and the MCQ with the correct options replaced by ``None of the above''.}
\label{figure:true-or-false} 
\end{figure*}

\section{Related Work}
LLMs \cite{gpt3,touvron2023llama2,chatgpt} have led the research of NLP into a new era. Recent advancements,
including supervised fine-tuning and alignment with human values \cite{instruct,chung2022scaling,bai2022training},
have further augmented the capabilities of LLMs, enabling them to adhere more closely to human instructions and ethical considerations.
Nonetheless, challenges persist since LLMs may show variability in the model responses, especially under the scenarios of MCQA.
\citet{robinson2022leveraging} termed the ability to associate the answer options and corresponding symbols as multiple choice symbol binding (MCSB)
and proved that the MCSB ability varied significantly by models. Additionally, \citet{wang2023large}
revealed vulnerabilities in the ranking of candidate responses, which could be manipulated by altering the presentation order. \citet{zheng2023large} investigated the token selection bias in LLMs. \citet{kadavath2022language} explored the reliability of the LLM performance and calibration, focusing exclusively on a set of private models under the MCQA settings. Recently, \citet{west2023generative} examined the performance gap between generative and discriminative tasks in LLMs. \citet{pezeshkpour2024large} proposed two calibration techniques to reduce variability in LLM responses. Previous work has focused on mitigating bias in answer options or developing techniques to ensure that LLMs exhibit consistency across different orders of options. A common thread among these studies is the belief that if LLMs can demonstrate robustness to variations in the order of answer options, their predictive reliability can be improved. However, our research identifies another limitation: even if LLMs consistently predict the correct answer across varied option orders, they may still struggle to accurately answer questions derived from the original MCQ because LLMs may perform MCQA by selecting which is the least incorrect.

%% file: Text/3.invariability_does_not.tex
\section{Does Invariability Imply Reliability?}
As discussed previously, prior research has demonstrated that LLMs may exhibit variability in their responses across different permutations of answer options in MCQs \cite{robinson2022leveraging,pezeshkpour2024large}. Various techniques have been explored to ensure that LLMs exhibit invariability in response to such permutations, with the assumption that invariability could serve as a proxy for model reliability in MCQ tasks. However, this raises an important question: does invariability truly equate to reliability?
\input{Table/true_or_false}
\input{Table/none_of_above}

\subsection{Models and Datasets}
In this study, we focus on evaluating several prominent generative models that have garnered significant attention within both academic and public domains. These include LLaMA models \cite{touvron2023llama2} (LLaMA 3 8B, LLaMA 2 13B\footnote{LLaMA 3 currently comprises models with 8B and 70B parameters}, LLaMA 3 70B), Mixtral~\cite{jiang2024mixtral} (Mixtral 8$\times$7B), and ChatGPT~\cite{chatgpt} (ChatGPT-3.5, ChatGPT-4o, and ChatGPT-4o-mini). For the datasets, we sample from two widely recognized benchmarks: the first is MMLU~\cite{mmlu}, a general-domain benchmark widely used in MCQA evaluation for LLMs; the second is MedMCQA~\cite{medmcqa}, which is specific to the medical domain and requires extensive domain-specific knowledge, presenting a significant challenge for most LLMs.

\subsection{Invariability Dataset Preparation}
Due to our work aiming to demonstrate certain deficiencies in using MCQs to test LLMs, and to facilitate subsequent experiments, we first conduct tests on MCQs with option permutations on subsets of MMLU and MedMCQA datasets with questions testing for knowledge instead of reasoning (like math). Then, we filter out the subsets MMLU$^\blacklozenge$ and MedMCQA$^\blacklozenge$ where the LLMs show invariability.

\subsection{Transforming to True-or-False Format}
\label{section:true_or_false}
We transform the original MCQAs in only MMLU$^\blacklozenge$ and MedMCQA$^\blacklozenge$ into a True-or-False (T/F) format to
explore how the LLMs behave on the questions that they have predicted correctly with invariability
in MCQA-format. For every MCQA instance, we generate T/F-format questions, including one
with the correct option (T/F: correct) and other questions with the incorrect options (T/F: incorrect) \footnote{``None of the above''-like options are not transformed into T/F format.}
as depicted in Figure \ref{figure:true-or-false}, anticipating that the LLMs will respond accurately
with ``Yes'' and ``No'' respectively.

Table~\ref{table:true_or_false} presents an analysis of LLM performance on T/F questions.
If consistency were a reliable indicator of accuracy, we would expect LLMs to achieve near-perfect
performance in this format on both the ``T/F: correct'' and ``T/F: incorrect'' datasets.
In the few-shot scenario, we provide the LLMs with two examples as demonstrations, one with the answer ``correct''
and the other with that of ``incorrect''.
In practice, LLMs demonstrate varying levels of accuracy on
the ``True/False: correct'' datasets, ranging from 70.3\% to 96.4\%. This suggests that LLMs can
generally perform well on T/F questions derived from MCQs with correct options.
However, a notable performance decline occurs when the T/F questions include incorrect
options. For instance, LLaMA 3 70B achieves an accuracy as low as 36.8\% on the ``T/F: incorrect''
datasets based on MedMCQA$^\blacklozenge$. Similar trends are observed across almost all tested
LLMs. This highlights a critical limitation: while LLMs tend to be consistent when handling
questions with both MCQA and T/F formats with correct options, they frequently
misclassify statements containing incorrect options as correct.

\begin{figure*}[t]
\centering
\includegraphics[width=0.99\textwidth]{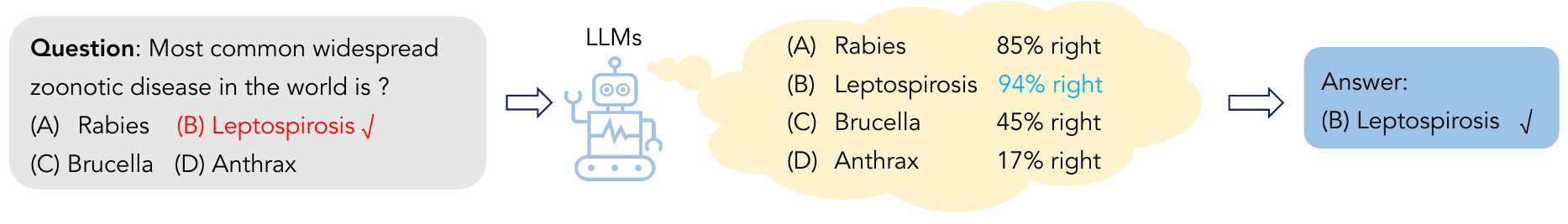}
\caption{Illustration of the hypothesis: LLMs may perform MCQA by selecting the least incorrect option.}
\label{figure:less_incorrect}
\end{figure*}

\begin{figure*}[htbp]
\centering
\includegraphics[width=0.99\textwidth]{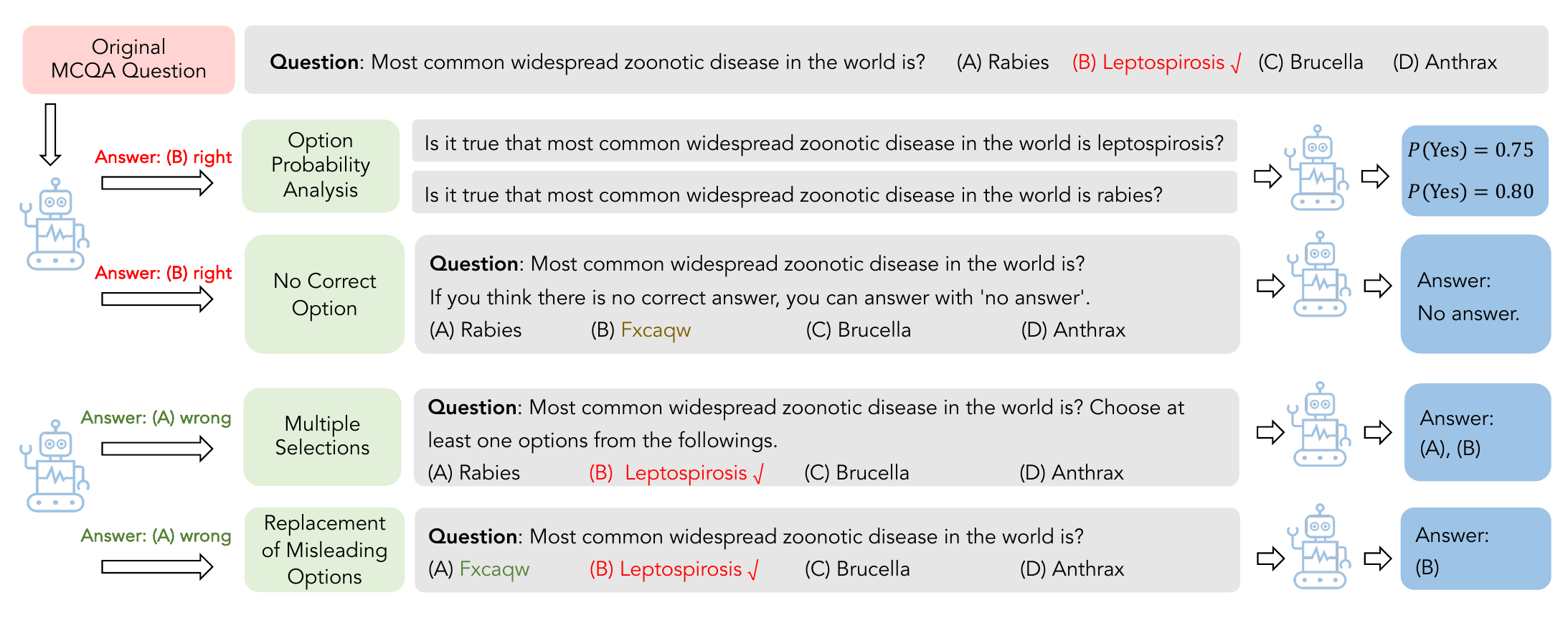}
\caption{The validation experiments consist of (1) ``Option probability analysis'' (2) ``No correct option'' on the MCQA$^\blacklozenge$, along with
(3) ``Multiple selections'' and (4) ``Replacement of misleading options'' on the original MCQs where
the LLMs made incorrect predictions. (1) assesses the confidence of LLMs when handling T/F questions with correct and incorrect answer options from the MCQA$^\blacklozenge$ datasets;
(2) assesses whether LLMs can respond with ``no answer'' when presented with MCQs that do not contain correct options;
(3) prompts the LLMs to selected all the answers options they consider correct in the MCQs where they previously made incorrect prediction.
(4) replaces the incorrect options previously chosen by the model with non-semantic tokens.}
\label{figure:three-exp}
\end{figure*}

\subsection{``None of the Above'' Options}
\citet{kadavath2022language} examined the potential impact of ``None of the above'' options
on certain close-source LLMs with the \textit{entire} MCQA datasets, without considering the confounding
factor of variability. In this study, we extend the analysis to a broader range of LLMs, focusing exclusively
on variability-free sub-datasets, that are MMLU$^\blacklozenge$ and MedMCQA$^\blacklozenge$, as illustrated in Figure~\ref{figure:true-or-false}.
For the few-shot scenarios, demonstrations involve MCQs with the correct answer of ``None of the above''.
As presented in Table~\ref{table:none}, the substitution of correct options with the ``None of the above''
leads to a substantial decline in model performance. With the exception of scenarios
involving ChatGPT-4o and ChatGPT-4o-mini, the LLMs consistently fail to select the ``None of the Above''
option in place of the correct answer, with accuracy not exceeding 48.5\% across all other scenarios.
Even the ChatGPT-4o model demonstrates a failure rate exceeding 31\% on the MCQA$^\blacklozenge$ datasets.

\subsection{Analysis}
Despite the invariability in LLM performance on MCQs, modifying the MCQA datasets to include
(1) T/F questions derived from incorrect options in the original MCQs,
and (2) scenarios where the correct option is replaced by ``None of the Above,'' results in a
pronounced performance decline. This finding highlights a critical issue: the invariability exhibited
by LLMs in handling multiple-choice questions does not necessarily signify reliability. Specifically,
the models demonstrate unexpected behavior when confronted with questions involving options other than the correct answer options,
raising concerns about the robustness of their decision-making processes in such contexts.

%% file: Table/true_or_false.tex
\begin{table*}[t]
\centering
\caption{Accuracy of LLMs on the True-or-False questions derived from MCQs.
$\blacklozenge$ means the subsets of datasets where LLMs have answered correctly across all re-ordered answer options of the MCQs.}
\label{table:true_or_false}
\begin{tabular}{l|l|cc|cc}
\hline
\multicolumn{1}{c|}{\multirow{2}{*}{}}        & \multirow{2}{*}{}   & \multicolumn{2}{c|}{0-shot}            & \multicolumn{2}{c}{few-shot}            \\ \cline{3-6}
\multicolumn{1}{c|}{}                         &                     & MMLU$^\blacklozenge$ & MedMCQA$^\blacklozenge$ & MMLU$^\blacklozenge$ & MedMCQA$^\blacklozenge$ \\ \hline
\multirow{2}{*}{LLaMA 3 8B}                   & \texttt{T/F: correct}      &  93.3                 & 92.4                &  95.1               &  93.0                      \\
                                              & \texttt{T/F: incorrect}     &  43.7         &   52.7                     &  48.9               &  55.5                  \\ \hline
\multirow{2}{*}{LLaMA 2 13B}                  & \texttt{T/F: correct} & 70.3                & 70.5                   & 73.1                & 71.8                   \\
                                              & \texttt{T/F: incorrect} & 52.5                & 55.8                   & 55.4                & 55.6                   \\ \hline
\multirow{2}{*}{LLaMA 3 70B}                  & \texttt{T/F: correct} &   97.2               &  92.3                       & 97.8                & 92.8                     \\
                                              & \texttt{T/F: incorrect} & 41.7                & 42.3                   &  44.1                & 36.8                        \\ \hline
\multirow{2}{*}{Mixtral 8$\times$7B}               & \texttt{T/F: correct} &   90.7             &  83.2                &    91.2                  &  82.7                       \\
                                              & \texttt{T/F: incorrect} & 58.8                &  54.7                     & 59.2                &  55.5                      \\ \hline
\multirow{2}{*}{ChatGPT-3.5}                  & \texttt{T/F: correct} & 87.6                & 76.6                   & 87.8                & 75.6                   \\
                                              & \texttt{T/F: incorrect} & 68.5                & 69.9                   & 68.9                & 71.6                   \\ \hline
\multirow{2}{*}{ChatGPT-4o-mini}              & \texttt{T/F: correct} &  93.6               &   89.0                  &  94.1               &  88.7                   \\
                                              & \texttt{T/F: incorrect} & 65.9                 &  72.4                  &  65.2               &  73.1                  \\ \hline
\multirow{2}{*}{ChatGPT-4o}                   & \texttt{T/F: correct} &    95.1             &  88.7                       &  96.4               &   89.9                 \\
                                              & \texttt{T/F: incorrect} &     73.0      &   80.2                       &   70.2                   &  80.4                    \\ \hline
\end{tabular}
\end{table*}

%% file: Table/none_of_above.tex
\begin{table*}[h]
\centering
\caption{Accuracy of LLMs on the MCQA$^\blacklozenge$ datasets with the correct options replaced with the ``None of the Above'' option.
$\blacklozenge$ denotes the subsets of datasets where LLMs have shown invariability across re-ordered answer options before the alteration of ``None of the above''.}
\label{table:none}
\vskip 0.1in
\begin{tabular}{l|cc|cc}
\hline
                      & \multicolumn{2}{c|}{0-shot}                                         & \multicolumn{2}{c}{few-shot}                                        \\ \cline{2-5}
\multicolumn{1}{c|}{} & \multicolumn{1}{c|}{MMLU$^\blacklozenge$} & MedMCQA$^\blacklozenge$ & \multicolumn{1}{c|}{MMLU$^\blacklozenge$} & MedMCQA$^\blacklozenge$ \\ \hline
LLaMA 3 8B             & \multicolumn{1}{c|}{18.3}             &  20.6              & \multicolumn{1}{c|}{19.1}        &  20.4                   \\
LLaMA 2 13B            & \multicolumn{1}{c|}{17.2}             &  12.3              & \multicolumn{1}{c|}{6.7}         &   0.0                      \\
LLaMA 3 70B            & \multicolumn{1}{c|}{22.6}             &  24.9              & \multicolumn{1}{c|}{31.3}        &  29.8                   \\ \hline
Mixtral 8$\times$7B   & \multicolumn{1}{c|}{24.7}             &  31.1              & \multicolumn{1}{c|}{40.3}        &  30.1                   \\ \hline
ChatGPT-3.5           & \multicolumn{1}{c|}{23.7}             &  36.0              & \multicolumn{1}{c|}{48.5}        &  41.7                       \\
ChatGPT-4o-mini       & \multicolumn{1}{c|}{43.6}             &  42.6              & \multicolumn{1}{c|}{50.6}        &  51.4                        \\
ChatGPT-4o            & \multicolumn{1}{c|}{60.3}             &  60.1              & \multicolumn{1}{c|}{68.6}        &  67.5                      \\ \hline
\end{tabular}
\end{table*}

%% file: Text/4.least_incorrect.tex
\input{Table/4-1-option}
\begin{table}[htbp]
\centering
\caption{Ratio of the instances where the LLMs can generate ``no answer'' on the MCQs with no correct option under the few-shot settings.}
\label{table:no_correct}
\resizebox{0.85\columnwidth}{!}{
\begin{tabular}{l|cc}
\hline
\multicolumn{1}{c|}{} & MMLU$^\blacklozenge$ & MedMCQA$^\blacklozenge$ \\ \hline
ChatGPT 4o-mini  & 32.8 & 32.6 \\
ChatGPT 4o       & 59.5 & 60.0\\ \hline
\end{tabular}
}
\end{table}

\section{LLMs May Do MCQA by Selecting Which Is the Least Incorrect}
\label{sec:least}
Through our experiments with modified datasets only with the incorrect options in the original MCQs,
we have shown that invariability in LLM responses does not necessarily equate to reliability.
Based on these observations, we propose the following hypothesis:

\textit{While LLMs demonstrate invariability on specific MCQs with a consistent answer option,
    they may not regard this option as uniquely correct. Rather, LLMs may treat the selected option
    as the most accurate among the choices, without dismissing the potential partial correctness of other,
    incorrect options—albeit to a lesser degree than the chosen one.}

\noindent which is visually illustrated in Figure~\ref{figure:less_incorrect}.

If this hypothesis holds true, it suggests that LLMs may recognize some of the unselected options
as partially correct. This could offer a plausible explanation for the observed model behavior
in the aforementioned experiments. To further investigate this phenomenon, we explore the behavior
of the models from the following four perspectives, as depicted in Figure~\ref{figure:three-exp}.

\subsection{Option Probability Analysis}
Leveraging the MMLU$^\blacklozenge$ and MedMCQA$^\blacklozenge$ datasets, we investigate the confidence of LLMs by examining
the distribution of token probabilities for answer option tokens, as discussed in~\citet{chen-etal-2023-close}.
To do this, we convert MCQs into T/F format as illustrated in Figure~\ref{figure:three-exp}.
The confidence scores for each answer option are derived based on the ``yes'' or ``no'' token probabilities assigned by the LLMs.
For T/F questions that include the correct answer options from the original MCQs,
we compute the confidence score $\text{C}_{correct}$ using instances where the LLMs made correct predictions.
This score quantifies the degree of confidence the LLMs exhibit when recognizing a claim with the correct option as accurate.
Conversely, for T/F questions containing incorrect answer options from the MCQs, we
compute the confidence score \(\text{C}_{incorrect}\) based on cases where the LLMs made incorrect predictions.
This score measures how confidently the LLMs mistakenly identify a claim with an incorrect option as correct.
For a specific MCQ, we consider the incorrect$^*$ option with the highest confidence in corresponding T/F questions for $\text{C}_{incorrect^*}$.

\begin{equation*}
    \text{C}_{correct} = \frac{1}{N} \sum_{z_{\text{'`yes''}}>z_{\text{``no''}}} \frac{e^{z_{\text{``yes''}}}}{\sum_{t \in V} e^{z_t}}
\end{equation*}

\begin{equation*}
    \text{C}_{incorrect} = \frac{1}{M} \sum_{z_{\text{``no''}}>z_{\text{``yes''}}} \frac{e^{z_{\text{``no''}}}}{\sum_{t \in V} e^{z_t}}
\end{equation*}

where \( z_t \) is the logit for each token \( t \) in the vocabulary, and \( V \) denotes
the full vocabulary set. \(N\) and \(M\) is the number of corresponding questions.

Table~\ref{table:prob} demonstrates the confidence of LLMs for the correct options ($\text{C}_{correct}$)
and the incorrect$^\ast$ options ($\text{C}_{incorrect^\ast}$), along with the relative confidence scores.
The experimental results show that while LLMs consistently consider the incorrect$^\ast$ options as
less correct than the correct options, demonstrated by all relative confidence being below 100\%,
the incorrect$^\ast$ options still achieve substantial confidence ranging from 83.6\% to 99.4\% of
those for the correct options. Consequently, despite invariability, LLMs may still perceive certain
incorrect options as correct, though to a lesser extent compared to the correct ones.

\input{Table/4-2-multi}

\subsection{MCQA with No Correct Option}
In our previously described scenarios for evaluating LLMs on MCQs, there has always been a correct answer among the candidate options.
However, when no correct answer is present, we expect LLMs to recognize that the question is flawed.
To guide the model in such cases, we prompt the LLMs with ``If you think there is no correct answer, you can respond with `no answer',''
to observe whether the model generates a ``no answer'' response.
Empirically, even large-scale open-source models such as LLaMA 3 70B struggle to effectively follow this instruction.
Consequently, our analysis focuses on the ChatGPT-4 series models.
Using the MCQA$^\blacklozenge$ dataset, where the models exhibit invariability, we replace the correct options with non-semantic tokens.
As shown in Table~\ref{table:no_correct}, with appropriately designed prompts, ChatGPT-4o successfully identifies that there is no correct answer in approximately 60\% of the MCQs.
However, for the remaining 40\% of the questions, it still selects one option as correct, indicating that LLMs do not fully recognize all incorrect options as incorrect.

\subsection{MCQA with Multiple Selections}
For the MCQA datasets involved in this study, LLMs are tasked with identifying only one correct
option per MCQ. We collect the instances where LLMs incorrectly predict the answers, denoted as
MMLU$^\dag$ and MedMCQA$^\dag$. In the above instances, the incorrect options which LLMs have
regarded as the correct ones mistakenly are defined as \textit{misleading} options. Then, LLMs are
prompted to recognize all plausible correct options among all answer options. Table~\ref{table:multiple_selection}
showcases the recall for the correct and misleading options for the instances
where LLMs render multiple selections. The results reveal that the correct options are included
in the selections in over 78.9\% of instances, reaching up to 94.2\%. This indicates that the LLMs also
recognize the correct options as correct but less correct than the misleading ones.

\subsection{MCQA with the Misleading Option Replacement}
Apart from the multi-selection scenario, we explore the impact of replacing misleading options with
arbitrary non-semantic tokens in MMLU$^\dag$ and MedMCQA$^\dag$. Table~\ref{table:replace_misleading}
elucidates that the LLMs correctly identify the correct options in 30.9\% to 58.0\% of instances,
highlighting the influence of misleading options on their predictions.
For cases where LLMs continue to make incorrect predictions, a fundamental deficit in relevant knowledge
likely underpins the LLM incapability to generate the correct answers.

\input{Table/4-3-replace}

\subsection{Summary}
The analyses conducted across the four experimental scenarios provide substantial support for the validity of the proposed hypothesis.
This leads to a critical observation that highlights a fundamental limitation of using MCQA-based evaluations to assess the capabilities of LLMs:
\textit{In the context of MCQA, while LLMs may select the correct answer, there remains a possibility that they also attribute correctness to other, incorrect options.}

%% file: Table/4-1-option.tex
\begin{table*}[tp]
\renewcommand{\arraystretch}{1.2}
\centering
\caption{Confidence of answer options in MCQA tasks with open-source LLMs. $\text{C}_{correct}$ : mean confidence of the correct options. $\text{C}_{incorrect^{\ast}}$: mean confidence of the incorrect options that score with the highest confidence. $R_c$: relative confidence score. $\blacklozenge$ denotes the experiments on the sub-datasets where the LLMs predict correctly with the original MCQA settings.}
\label{table:prob}
\resizebox{0.99\textwidth}{!}{
\begin{tabular}{l|cc|cc|cc|cc}
\hline
                          & \multicolumn{2}{c|}{LLaMA 3 8B}        & \multicolumn{2}{c|}{LLaMA 2 13B}       & \multicolumn{2}{c|}{LLaMA 3 70B}       & \multicolumn{2}{c}{Mixtral 8$\times$7B}                          \\ \cline{2-9}
                          & MMLU$^\blacklozenge$ & MedMCQA$^\blacklozenge$ & MMLU$^\blacklozenge$ & MedMCQA$^\blacklozenge$ & MMLU$^\blacklozenge$ & MedMCQA$^\blacklozenge$ & MMLU$^\blacklozenge$ & \multicolumn{1}{c}{MedMCQA$^\blacklozenge$} \\ \hline
$\text{C}_{correct}$        & 16.4      &  18.7        & 35.0     & 36.0                   &   23.6         &  25.5             &  31.2       &  34.6                                           \\
$\text{C}_{incorrect^{\ast}}$  & 16.3      &  18.5        & 32.5     & 30.1                   &   20.3         &  21.9             &  29.1       &  33.1                                       \\ \hline
$R_c$                       & 99.4\%    &  98.9\%      & 92.9\%   & 83.6\%                 &   86.0\%       &  85.9\%           &  93.3\%     &  95.7\%                                      \\ \hline \end{tabular}
}
\end{table*}

%% file: Table/4-2-multi.tex
\begin{table*}[htbpp]
\centering
\caption{Experiments on the altered MCQA datasets with multiple selections.
\texttt{Recall$_{correct}$}: recall of the correct options.
\texttt{Recall$_{misleading}$}: recall of the misleading options (the incorrect options LLMs have chosen).
$\dag$ denotes the subsets where the LLMs have generated incorrect answers on the original MCQA datasets.}
\label{table:multiple_selection}
\resizebox{0.85\textwidth}{!}{
\begin{tabular}{l|cc|cc}
\hline
\multicolumn{1}{c|}{} & \multicolumn{2}{c|}{MMLU$^\dag$}                                                & \multicolumn{2}{c}{MedMCQA$^\dag$}                                              \\ \cline{2-5}
\multicolumn{1}{c|}{} & \texttt{Recall}$_{correct}$ & \texttt{Recall}$_{misleading}$  & \texttt{Recall}$_{correct}$ & \texttt{Recall}$_{misleading}$ \\ \hline
LLaMA 3 8B             & 85.1              &  70.1          &  82.3         & 74.2                                   \\
LLaMA 2 13B            & 92.5              &  67.5          &  78.9         & 70.3                                  \\
LLaMA 3 70B            & 94.2              &  72.2          &  84.0         & 80.1                                   \\ \hline
Mixtral 8$\times$7B   & 91.5              &  76.4          &  84.6         & 85.4                                   \\ \hline
ChatGPT-3.5           & 85.1              &  66.7          &  81.3         & 69.7                           \\
ChatGPT-4o-mini       & 85.0              &  89.2          &  84.4         & 91.6                                   \\
ChatGPT-4o            & 89.9              &  90.8          &  83.2         & 92.1                                   \\ \hline
\end{tabular}
}
\end{table*}

%% file: Table/4-3-replace.tex
\begin{table}[htbp]
\renewcommand{\arraystretch}{1.1}
\centering
\caption{Ratio of the instances where the LLMs turn to predict correctly with the replacement of misleading options (the incorrect options LLMs have chosen).}
\label{table:replace_misleading}
\resizebox{0.85\columnwidth}{!}{
\begin{tabular}{l|cc}
\hline
\multicolumn{1}{c|}{}        & MMLU$^\dag$ & MedMCQA$^\dag$ \\ \hline
LLaMA 3 8B          & 42.1         & 30.9             \\
LLaMA 2 13B          & 58.0         & 41.0            \\
LLaMA 3 70B         & 46.3         & 41.1             \\ \hline
Mixtral 8$\times$7B      & 50.4         & 34.8             \\ \hline
ChatGPT-3.5         & 53.6         & 44.5            \\
ChatGPT-4o-mini          & 39.7         & 41.2             \\
ChatGPT-4o     & 41.1         & 48.6             \\ \hline
\end{tabular}
}
\end{table}

%% file: Text/6.MCQA_plus.tex
\input{Table/5-mcqa-plus}
\section{MCQA+ for Robust Evaluation}
\paragraph{Dataset Preparation}
Experimental analyses have revealed significant limitations in using the MCQA benchmark to evaluate LLMs,
highlighting that LLMs may consider options they did not select in MCQs as correct.
To address this, we propose an augmentation approach based on the original MCQA dataset,
informed by the empirical findings from the above experiments.
Each MCQ is transformed into one of the following settings:
(a) Original MCQs;
(b) MCQs with re-ordered answer options;
(c) True-or-False questions derived from correct answer options;
(d) True-or-False questions derived from incorrect answer options;
(e) MCQs where the correct options are replaced with ``None of the above''; and
(f) MCQs with no correct options, where LLMs are expected to generate ``no answer'' as the response.
Using these settings, we propose three dataset augmentation approaches:
(1) \textbf{MCQA+}: Encompasses all of the above settings;
(2) \textbf{MCQA+$^{hard}$}: Includes only the settings (b, d, e, f), serving as a much more challenging benchmark for LLMs.
(3) \textbf{MCQA+ ($\times 1$)}: Samples one question from the MCQA+ settings as an efficient approximation to MCQA+.

The mean accuracy across all settings is adopted as the evaluation metric for LLMs.
For settings with multiple questions (e.g., (b)), $\text{accuracy}_b$ is measured as the mean accuracy across all questions in setting (b).
Table~\ref{table:mcqa-plus} illustrates the comparative performance of LLMs on the original MCQA dataset, MCQA+, MCQA+$^{hard}$, and MCQA+ ($\times 1$).
Performance on the MCQA+ dataset shows a significant decline across all LLMs compared to the original MCQA dataset.
For example, accuracy for LLaMA 3 8B dropped from 75.1\% to 56.8\% on the MMLU dataset.
Performance on the MCQA+$^{hard}$ benchmark is even lower, likely for reasons discussed in Section~\ref{sec:least}.
Even ChatGPT-4o experienced a performance decline of 9.3\% from the original MCQA to MCQA+$^{hard}$ on the MMLU dataset.

Although MCQA+ and MCQA+$^{hard}$ provide a more accurate reflection of LLM capabilities,
they entail significantly higher computational costs compared to the original MCQA.
Therefore, MCQA+ ($\times 1$), which samples from MCQA+ for each MCQ, requires no additional computational cost compared to the original MCQA.
As shown in Table~\ref{table:mcqa-plus}, this cost-efficient approach still effectively reveals the true capabilities of LLMs.

\paragraph{Discussion}
The MCQA+ strategy offers an efficient and refined approach to augmenting existing MCQ datasets,
enabling a more accurate assessment of model capability.
However, ensuring consistent performance across tasks not addressed by MCQA+, such as generative tasks, remains a challenge.
Based on the results of this study, we hypothesize that the observed performance decline may be linked to the training strategies of LLMs
in generative tasks during pre-training and instruction-tuning, that is predicting the next token based on the ranking of probability.
LLMs have been only instructed to choose the best options but not to treat those options as exclusively correct.
While reinforcement learning aligns the model's outputs with human preferences,
it does not fully resolve the issue where incorrect options might still receive high probabilities in different contexts.
This could explain the discrepancies in model performance between discriminative and generative tasks, as noted by \citet{west2023generative}.
As such, the reliability of evaluating LLMs using MCQs necessitates further scrutiny and attention.

%% file: Table/5-mcqa-plus.tex
\begin{table*}[htb]
\renewcommand{\arraystretch}{1.1}
\centering
\caption{Model performance on the original MCQA, MCQA+, MCQA+$^{hard}$ and MCQA+ ($\times$1) datasets.}
\label{table:mcqa-plus}
\resizebox{0.98\textwidth}{!}{
\begin{tabular}{l|cccc|cccc}
\hline
\multicolumn{1}{c|}{} & \multicolumn{4}{c|}{MMLU}                                   & \multicolumn{4}{c}{MedMCQA}                                 \\ \cline{2-9}
                      & MCQA & MCQA+ & MCQA+$^{hard}$ & MCQA+ ($\times$1) & MCQA & MCQA+ & MCQA+$^{hard}$ & MCQA+ ($\times$1) \\ \hline
LLaMA 3 8B            & 75.1 & 56.8  & 40.5  & 58.4    & 47.9 & 36.3  & 24.4   & 34.1                  \\
LLaMA 2 13B            & 72.7 & 46.1  & 21.2  & 45.3    & 43.2 & 38.8  & 16.5   & 40.0              \\
LLaMA 3 70B           & 78.9 & 60.2  & 46.8  & 57.1    & 53.1 & 42.8  & 29.1   & 44.4              \\ \hline
Mixtral 8$\times$7B   & 71.2 & 58.6  & 43.7  & 58.5    & 51.4 & 43.5  & 28.7   & 42.7              \\ \hline
ChatGPT-3.5           & 65.0 & 63.2  & 57.8  & 64.0    & 56.9 & 53.9  & 49.6   & 54.1              \\
ChatGPT-4o-mini       & 79.0 & 70.9  & 63.0  & 72.4    & 68.3 & 64.4  & 60.2   & 63.8                  \\
ChatGPT-4o            & 82.4 & 80.7  & 73.1  & 79.7    & 72.7 & 69.7  & 64.0   & 71.3               \\ \hline
\end{tabular}
}
\end{table*}

%% file: Text/7.conclusion.tex
\section{Conclusion}
In this study, we investigated the limitations of using MCQA as a benchmark for evaluating the performance of LLMs through a comprehensive series of experiments.
Our findings suggest that LLMs may not always select the distinctly correct option, but instead opt for the least incorrect option.
This behavior raises concerns about the robustness and reliability of MCQA-based evaluations.
To address these issues, we proposed the MCQA+ dataset augmentation method, which provides a more refined evaluation
framework by challenging LLMs to demonstrate a deeper level of understanding.
Our work underscores the importance of continued efforts to develop more comprehensive evaluation
methodologies for LLMs, ensuring that their true capabilities are accurately reflected,
not only in discriminative tasks, such as MCQA but also in broader, more complex contexts.

%% file: Text/8.limitation_n_ethics.tex
\section*{Limitations}
In this study, we analyzed a probable issue that LLMs may face when answering MCQs.
Building on previous research on the variability of large models, we conducted experiments
demonstrating that although LLMs can achieve impressive results on MCQA benchmarks,
their treatment of incorrect options may be ambiguous, potentially recognizing incorrect options in other contexts.
This issue may stem from negative impacts introduced by different stages of the training objectives of LLMs,
such as instruction-tuning and RL-based alignment, presenting a broader challenge for the entire NLP field.
Therefore, we propose a method to improve the reliability of model evaluations through diversity testing,
which represents a trade-off between efficiency and accuracy, without fundamentally addressing the
core challenges of evaluating LLMs based on MCQs.
We aim to draw attention from the community to the potential long-term impacts of this issue and to
collaboratively work towards resolving it.

%% file: Text/8.z_acknowledgement.tex
\section*{Acknowledgements}
We thank the anonymous reviewers for their insightful and constructive comments and gratefully acknowledge
the support of the National Natural Science Foundation of China [62206079]; and the Heilongjiang
Provincial Natural Science Foundation of China [2023ZX01A11].
We also appreciate the support from China Mobile Group Heilongjiang Co., Ltd. @ on our research, the research is jointly completed by both parties.

%% file: coling_mcqa.bbl
\begin{thebibliography}{30}
\providecommand{\natexlab}[1]{#1}

\bibitem[{Bai et~al.(2022)Bai, Jones, Ndousse, Askell, Chen, DasSarma, Drain,
  Fort, Ganguli, Henighan et~al.}]{bai2022training}
Yuntao Bai, Andy Jones, Kamal Ndousse, Amanda Askell, Anna Chen, Nova DasSarma,
  Dawn Drain, Stanislav Fort, Deep Ganguli, Tom Henighan, et~al. 2022.
\newblock Training a helpful and harmless assistant with reinforcement learning
  from human feedback.
\newblock \emph{arXiv preprint arXiv:2204.05862}.

\bibitem[{Brown et~al.(2020)Brown, Mann, Ryder, Subbiah, Kaplan, Dhariwal,
  Neelakantan, Shyam, Sastry, Askell, Agarwal, Herbert{-}Voss, Krueger,
  Henighan, Child, Ramesh, Ziegler, Wu, Winter, Hesse, Chen, Sigler, Litwin,
  Gray, Chess, Clark, Berner, McCandlish, Radford, Sutskever, and
  Amodei}]{gpt3}
Tom~B. Brown, Benjamin Mann, Nick Ryder, Melanie Subbiah, Jared Kaplan,
  Prafulla Dhariwal, Arvind Neelakantan, Pranav Shyam, Girish Sastry, Amanda
  Askell, Sandhini Agarwal, Ariel Herbert{-}Voss, Gretchen Krueger, Tom
  Henighan, Rewon Child, Aditya Ramesh, Daniel~M. Ziegler, Jeffrey Wu, Clemens
  Winter, Christopher Hesse, Mark Chen, Eric Sigler, Mateusz Litwin, Scott
  Gray, Benjamin Chess, Jack Clark, Christopher Berner, Sam McCandlish, Alec
  Radford, Ilya Sutskever, and Dario Amodei. 2020.
\newblock \href {https://arxiv.org/abs/2005.14165} {Language models are
  few-shot learners}.
\newblock \emph{CoRR}, abs/2005.14165.

\bibitem[{Chang et~al.(2023)Chang, Wang, Wang, Wu, Zhu, Chen, Yang, Yi, Wang,
  Wang et~al.}]{chang2023survey}
Yupeng Chang, Xu~Wang, Jindong Wang, Yuan Wu, Kaijie Zhu, Hao Chen, Linyi Yang,
  Xiaoyuan Yi, Cunxiang Wang, Yidong Wang, et~al. 2023.
\newblock A survey on evaluation of large language models.
\newblock \emph{arXiv preprint arXiv:2307.03109}.

\bibitem[{Chen et~al.(2023)Chen, Yuan, Cui, Liu, and Ji}]{chen-etal-2023-close}
Yangyi Chen, Lifan Yuan, Ganqu Cui, Zhiyuan Liu, and Heng Ji. 2023.
\newblock A close look into the calibration of pre-trained language models.
\newblock In \emph{Proceedings of the 61st Annual Meeting of the Association
  for Computational Linguistics (Volume 1: Long Papers)}, pages 1343--1367,
  Toronto, Canada. Association for Computational Linguistics.

\bibitem[{Chung et~al.(2022)Chung, Hou, Longpre, Zoph, Tay, Fedus, Li, Wang,
  Dehghani, Brahma et~al.}]{chung2022scaling}
Hyung~Won Chung, Le~Hou, Shayne Longpre, Barret Zoph, Yi~Tay, William Fedus,
  Yunxuan Li, Xuezhi Wang, Mostafa Dehghani, Siddhartha Brahma, et~al. 2022.
\newblock Scaling instruction-finetuned language models.
\newblock \emph{arXiv preprint arXiv:2210.11416}.

\bibitem[{Gao et~al.(2021)Gao, Tow, Biderman, Black, DiPofi, Foster, Golding,
  Hsu, McDonell, Muennighoff, Phang, Reynolds, Tang, Thite, Wang, Wang, and
  Zou}]{leo_gao_2021_5371629}
Leo Gao, Jonathan Tow, Stella Biderman, Sid Black, Anthony DiPofi, Charles
  Foster, Laurence Golding, Jeffrey Hsu, Kyle McDonell, Niklas Muennighoff,
  Jason Phang, Laria Reynolds, Eric Tang, Anish Thite, Ben Wang, Kevin Wang,
  and Andy Zou. 2021.
\newblock \href {https://doi.org/10.5281/zenodo.5371629} {A framework for
  few-shot language model evaluation}.

\bibitem[{Hendrycks et~al.(2021)Hendrycks, Burns, Basart, Zou, Mazeika, Song,
  and Steinhardt}]{mmlu}
Dan Hendrycks, Collin Burns, Steven Basart, Andy Zou, Mantas Mazeika, Dawn
  Song, and Jacob Steinhardt. 2021.
\newblock Measuring massive multitask language understanding.
\newblock \emph{Proceedings of the International Conference on Learning
  Representations (ICLR)}.

\bibitem[{Huang et~al.(2019)Huang, Le~Bras, Bhagavatula, and Choi}]{cqa}
Lifu Huang, Ronan Le~Bras, Chandra Bhagavatula, and Yejin Choi. 2019.
\newblock Cosmos {QA}: Machine reading comprehension with contextual
  commonsense reasoning.
\newblock In \emph{Proceedings of the 2019 Conference on Empirical Methods in
  Natural Language Processing and the 9th International Joint Conference on
  Natural Language Processing (EMNLP-IJCNLP)}, pages 2391--2401, Hong Kong,
  China. Association for Computational Linguistics.

\bibitem[{Jiang et~al.(2024)Jiang, Sablayrolles, Roux, Mensch, Savary, Bamford,
  Chaplot, Casas, Hanna, Bressand et~al.}]{jiang2024mixtral}
Albert~Q Jiang, Alexandre Sablayrolles, Antoine Roux, Arthur Mensch, Blanche
  Savary, Chris Bamford, Devendra~Singh Chaplot, Diego de~las Casas, Emma~Bou
  Hanna, Florian Bressand, et~al. 2024.
\newblock Mixtral of experts.
\newblock \emph{arXiv preprint arXiv:2401.04088}.

\bibitem[{Kadavath et~al.(2022)Kadavath, Conerly, Askell, Henighan, Drain,
  Perez, Schiefer, Dodds, DasSarma, Tran-Johnson et~al.}]{kadavath2022language}
Saurav Kadavath, Tom Conerly, Amanda Askell, Tom Henighan, Dawn Drain, Ethan
  Perez, Nicholas Schiefer, Zac~Hatfield Dodds, Nova DasSarma, Eli
  Tran-Johnson, et~al. 2022.
\newblock Language models (mostly) know what they know.
\newblock \emph{arXiv preprint arXiv:2207.05221}.

\bibitem[{Lai et~al.(2017)Lai, Xie, Liu, Yang, and Hovy}]{race}
Guokun Lai, Qizhe Xie, Hanxiao Liu, Yiming Yang, and Eduard Hovy. 2017.
\newblock {RACE}: Large-scale {R}e{A}ding comprehension dataset from
  examinations.
\newblock In \emph{Proceedings of the 2017 Conference on Empirical Methods in
  Natural Language Processing}, pages 785--794, Copenhagen, Denmark.
  Association for Computational Linguistics.

\bibitem[{Lin(2004)}]{lin-2004-rouge}
Chin-Yew Lin. 2004.
\newblock {ROUGE}: A package for automatic evaluation of summaries.
\newblock In \emph{Text Summarization Branches Out}, pages 74--81, Barcelona,
  Spain. Association for Computational Linguistics.

\bibitem[{Mostafazadeh et~al.(2016)Mostafazadeh, Chambers, He, Parikh, Batra,
  Vanderwende, Kohli, and Allen}]{sc}
Nasrin Mostafazadeh, Nathanael Chambers, Xiaodong He, Devi Parikh, Dhruv Batra,
  Lucy Vanderwende, Pushmeet Kohli, and James Allen. 2016.
\newblock A corpus and cloze evaluation for deeper understanding of commonsense
  stories.
\newblock In \emph{Proceedings of the 2016 Conference of the North {A}merican
  Chapter of the Association for Computational Linguistics: Human Language
  Technologies}. Association for Computational Linguistics.

\bibitem[{OpenAI et~al.(2023)OpenAI, :, Achiam, Adler, Agarwal, Ahmad, Akkaya,
  Aleman, Almeida, Altenschmidt, Altman, Anadkat, Avila, Babuschkin, Balaji,
  Balcom, Baltescu, Bao, Bavarian, Belgum, Bello, Berdine, Bernadett-Shapiro,
  Berner, Bogdonoff, Boiko, Boyd, Brakman, Brockman, Brooks, Brundage, Button,
  Cai, Campbell, Cann, Carey, Carlson, Carmichael, Chan, Chang, Chantzis, Chen,
  Chen, Chen, Chen, Chen, Chess, Cho, Chu, Chung, Cummings, Currier, Dai,
  Decareaux, Degry, Deutsch, Deville, Dhar, Dohan, Dowling, Dunning, Ecoffet,
  Eleti, Eloundou, Farhi, Fedus, Felix, Fishman, Forte, Fulford, Gao, Georges,
  Gibson, Goel, Gogineni, Goh, Gontijo-Lopes, Gordon, Grafstein, Gray, Greene,
  Gross, Gu, Guo, Hallacy, Han, Harris, He, Heaton, Heidecke, Hesse, Hickey,
  Hickey, Hoeschele, Houghton, Hsu, Hu, Hu, Huizinga, Jain, Jain, Jang, Jiang,
  Jiang, Jin, Jin, Jomoto, Jonn, Jun, Kaftan, Łukasz Kaiser, Kamali,
  Kanitscheider, Keskar, Khan, Kilpatrick, Kim, Kim, Kim, Kirchner, Kiros,
  Knight, Kokotajlo, Łukasz Kondraciuk, Kondrich, Konstantinidis, Kosic,
  Krueger, Kuo, Lampe, Lan, Lee, Leike, Leung, Levy, Li, Lim, Lin, Lin, Litwin,
  Lopez, Lowe, Lue, Makanju, Malfacini, Manning, Markov, Markovski, Martin,
  Mayer, Mayne, McGrew, McKinney, McLeavey, McMillan, McNeil, Medina, Mehta,
  Menick, Metz, Mishchenko, Mishkin, Monaco, Morikawa, Mossing, Mu, Murati,
  Murk, Mély, Nair, Nakano, Nayak, Neelakantan, Ngo, Noh, Ouyang, O'Keefe,
  Pachocki, Paino, Palermo, Pantuliano, Parascandolo, Parish, Parparita,
  Passos, Pavlov, Peng, Perelman, de~Avila Belbute~Peres, Petrov,
  de~Oliveira~Pinto, Michael, Pokorny, Pokrass, Pong, Powell, Power, Power,
  Proehl, Puri, Radford, Rae, Ramesh, Raymond, Real, Rimbach, Ross, Rotsted,
  Roussez, Ryder, Saltarelli, Sanders, Santurkar, Sastry, Schmidt, Schnurr,
  Schulman, Selsam, Sheppard, Sherbakov, Shieh, Shoker, Shyam, Sidor, Sigler,
  Simens, Sitkin, Slama, Sohl, Sokolowsky, Song, Staudacher, Such, Summers,
  Sutskever, Tang, Tezak, Thompson, Tillet, Tootoonchian, Tseng, Tuggle,
  Turley, Tworek, Uribe, Vallone, Vijayvergiya, Voss, Wainwright, Wang, Wang,
  Wang, Ward, Wei, Weinmann, Welihinda, Welinder, Weng, Weng, Wiethoff,
  Willner, Winter, Wolrich, Wong, Workman, Wu, Wu, Wu, Xiao, Xu, Yoo, Yu, Yuan,
  Zaremba, Zellers, Zhang, Zhang, Zhao, Zheng, Zhuang, Zhuk, and
  Zoph}]{openai2023gpt4}
OpenAI, :, Josh Achiam, Steven Adler, Sandhini Agarwal, Lama Ahmad, Ilge
  Akkaya, Florencia~Leoni Aleman, Diogo Almeida, Janko Altenschmidt, Sam
  Altman, Shyamal Anadkat, Red Avila, Igor Babuschkin, Suchir Balaji, Valerie
  Balcom, Paul Baltescu, Haiming Bao, Mo~Bavarian, Jeff Belgum, Irwan Bello,
  Jake Berdine, Gabriel Bernadett-Shapiro, Christopher Berner, Lenny Bogdonoff,
  Oleg Boiko, Madelaine Boyd, Anna-Luisa Brakman, Greg Brockman, Tim Brooks,
  Miles Brundage, Kevin Button, Trevor Cai, Rosie Campbell, Andrew Cann,
  Brittany Carey, Chelsea Carlson, Rory Carmichael, Brooke Chan, Che Chang,
  Fotis Chantzis, Derek Chen, Sully Chen, Ruby Chen, Jason Chen, Mark Chen, Ben
  Chess, Chester Cho, Casey Chu, Hyung~Won Chung, Dave Cummings, Jeremiah
  Currier, Yunxing Dai, Cory Decareaux, Thomas Degry, Noah Deutsch, Damien
  Deville, Arka Dhar, David Dohan, Steve Dowling, Sheila Dunning, Adrien
  Ecoffet, Atty Eleti, Tyna Eloundou, David Farhi, Liam Fedus, Niko Felix,
  Simón~Posada Fishman, Juston Forte, Isabella Fulford, Leo Gao, Elie Georges,
  Christian Gibson, Vik Goel, Tarun Gogineni, Gabriel Goh, Rapha Gontijo-Lopes,
  Jonathan Gordon, Morgan Grafstein, Scott Gray, Ryan Greene, Joshua Gross,
  Shixiang~Shane Gu, Yufei Guo, Chris Hallacy, Jesse Han, Jeff Harris, Yuchen
  He, Mike Heaton, Johannes Heidecke, Chris Hesse, Alan Hickey, Wade Hickey,
  Peter Hoeschele, Brandon Houghton, Kenny Hsu, Shengli Hu, Xin Hu, Joost
  Huizinga, Shantanu Jain, Shawn Jain, Joanne Jang, Angela Jiang, Roger Jiang,
  Haozhun Jin, Denny Jin, Shino Jomoto, Billie Jonn, Heewoo Jun, Tomer Kaftan,
  Łukasz Kaiser, Ali Kamali, Ingmar Kanitscheider, Nitish~Shirish Keskar,
  Tabarak Khan, Logan Kilpatrick, Jong~Wook Kim, Christina Kim, Yongjik Kim,
  Hendrik Kirchner, Jamie Kiros, Matt Knight, Daniel Kokotajlo, Łukasz
  Kondraciuk, Andrew Kondrich, Aris Konstantinidis, Kyle Kosic, Gretchen
  Krueger, Vishal Kuo, Michael Lampe, Ikai Lan, Teddy Lee, Jan Leike, Jade
  Leung, Daniel Levy, Chak~Ming Li, Rachel Lim, Molly Lin, Stephanie Lin,
  Mateusz Litwin, Theresa Lopez, Ryan Lowe, Patricia Lue, Anna Makanju, Kim
  Malfacini, Sam Manning, Todor Markov, Yaniv Markovski, Bianca Martin, Katie
  Mayer, Andrew Mayne, Bob McGrew, Scott~Mayer McKinney, Christine McLeavey,
  Paul McMillan, Jake McNeil, David Medina, Aalok Mehta, Jacob Menick, Luke
  Metz, Andrey Mishchenko, Pamela Mishkin, Vinnie Monaco, Evan Morikawa, Daniel
  Mossing, Tong Mu, Mira Murati, Oleg Murk, David Mély, Ashvin Nair, Reiichiro
  Nakano, Rajeev Nayak, Arvind Neelakantan, Richard Ngo, Hyeonwoo Noh, Long
  Ouyang, Cullen O'Keefe, Jakub Pachocki, Alex Paino, Joe Palermo, Ashley
  Pantuliano, Giambattista Parascandolo, Joel Parish, Emy Parparita, Alex
  Passos, Mikhail Pavlov, Andrew Peng, Adam Perelman, Filipe de~Avila
  Belbute~Peres, Michael Petrov, Henrique~Ponde de~Oliveira~Pinto, Michael,
  Pokorny, Michelle Pokrass, Vitchyr Pong, Tolly Powell, Alethea Power, Boris
  Power, Elizabeth Proehl, Raul Puri, Alec Radford, Jack Rae, Aditya Ramesh,
  Cameron Raymond, Francis Real, Kendra Rimbach, Carl Ross, Bob Rotsted, Henri
  Roussez, Nick Ryder, Mario Saltarelli, Ted Sanders, Shibani Santurkar, Girish
  Sastry, Heather Schmidt, David Schnurr, John Schulman, Daniel Selsam, Kyla
  Sheppard, Toki Sherbakov, Jessica Shieh, Sarah Shoker, Pranav Shyam, Szymon
  Sidor, Eric Sigler, Maddie Simens, Jordan Sitkin, Katarina Slama, Ian Sohl,
  Benjamin Sokolowsky, Yang Song, Natalie Staudacher, Felipe~Petroski Such,
  Natalie Summers, Ilya Sutskever, Jie Tang, Nikolas Tezak, Madeleine Thompson,
  Phil Tillet, Amin Tootoonchian, Elizabeth Tseng, Preston Tuggle, Nick Turley,
  Jerry Tworek, Juan Felipe~Cerón Uribe, Andrea Vallone, Arun Vijayvergiya,
  Chelsea Voss, Carroll Wainwright, Justin~Jay Wang, Alvin Wang, Ben Wang,
  Jonathan Ward, Jason Wei, CJ~Weinmann, Akila Welihinda, Peter Welinder, Jiayi
  Weng, Lilian Weng, Matt Wiethoff, Dave Willner, Clemens Winter, Samuel
  Wolrich, Hannah Wong, Lauren Workman, Sherwin Wu, Jeff Wu, Michael Wu, Kai
  Xiao, Tao Xu, Sarah Yoo, Kevin Yu, Qiming Yuan, Wojciech Zaremba, Rowan
  Zellers, Chong Zhang, Marvin Zhang, Shengjia Zhao, Tianhao Zheng, Juntang
  Zhuang, William Zhuk, and Barret Zoph. 2023.
\newblock \href {https://arxiv.org/abs/2303.08774} {Gpt-4 technical report}.
\newblock \emph{Preprint}, arXiv:2303.08774.

\bibitem[{OpenAI(2022)}]{chatgpt}
OpenAI. 2022.
\newblock Chatgpt.
\newblock \url{https://chat.openai.com}.

\bibitem[{Ouyang et~al.(2022)Ouyang, Wu, Jiang, Almeida, Wainwright, Mishkin,
  Zhang, Agarwal, Slama, Ray, Schulman, Hilton, Kelton, Miller, Simens, Askell,
  Welinder, Christiano, Leike, and Lowe}]{instruct}
Long Ouyang, Jeff Wu, Xu~Jiang, Diogo Almeida, Carroll~L. Wainwright, Pamela
  Mishkin, Chong Zhang, Sandhini Agarwal, Katarina Slama, Alex Ray, John
  Schulman, Jacob Hilton, Fraser Kelton, Luke Miller, Maddie Simens, Amanda
  Askell, Peter Welinder, Paul Christiano, Jan Leike, and Ryan Lowe. 2022.
\newblock Training language models to follow instructions with human feedback.
\newblock \emph{arXiv preprint arXiv:2203.02155}.

\bibitem[{Pal et~al.(2022)Pal, Umapathi, and Sankarasubbu}]{medmcqa}
Ankit Pal, Logesh~Kumar Umapathi, and Malaikannan Sankarasubbu. 2022.
\newblock Medmcqa: A large-scale multi-subject multi-choice dataset for medical
  domain question answering.
\newblock In \emph{Proceedings of the Conference on Health, Inference, and
  Learning}, volume 174 of \emph{Proceedings of Machine Learning Research},
  pages 248--260. PMLR.

\bibitem[{Papineni et~al.(2002)Papineni, Roukos, Ward, and
  Zhu}]{papineni-etal-2002-bleu}
Kishore Papineni, Salim Roukos, Todd Ward, and Wei-Jing Zhu. 2002.
\newblock {B}leu: a method for automatic evaluation of machine translation.
\newblock In \emph{Proceedings of the 40th Annual Meeting of the Association
  for Computational Linguistics}, pages 311--318, Philadelphia, Pennsylvania,
  USA. Association for Computational Linguistics.

\bibitem[{Pezeshkpour and Hruschka(2024)}]{pezeshkpour2024large}
Pouya Pezeshkpour and Estevam Hruschka. 2024.
\newblock Large language models sensitivity to the order of options in
  multiple-choice questions.
\newblock In \emph{Findings of the Association for Computational Linguistics:
  NAACL 2024}, pages 2006--2017.

\bibitem[{Robinson and Wingate(2022)}]{robinson2022leveraging}
Joshua Robinson and David Wingate. 2022.
\newblock Leveraging large language models for multiple choice question
  answering.
\newblock In \emph{The Eleventh International Conference on Learning
  Representations}.

\bibitem[{Sap et~al.(2019)Sap, Rashkin, Chen, Le~Bras, and Choi}]{siqa}
Maarten Sap, Hannah Rashkin, Derek Chen, Ronan Le~Bras, and Yejin Choi. 2019.
\newblock Social {IQ}a: Commonsense reasoning about social interactions.
\newblock In \emph{Proceedings of the 2019 Conference on Empirical Methods in
  Natural Language Processing and the 9th International Joint Conference on
  Natural Language Processing (EMNLP-IJCNLP)}, pages 4463--4473, Hong Kong,
  China. Association for Computational Linguistics.

\bibitem[{Talmor et~al.(2019)Talmor, Herzig, Lourie, and Berant}]{csqa}
Alon Talmor, Jonathan Herzig, Nicholas Lourie, and Jonathan Berant. 2019.
\newblock {C}ommonsense{QA}: A question answering challenge targeting
  commonsense knowledge.
\newblock In \emph{Proceedings of the 2019 Conference of the North {A}merican
  Chapter of the Association for Computational Linguistics: Human Language
  Technologies, Volume 1 (Long and Short Papers)}, pages 4149--4158,
  Minneapolis, Minnesota. Association for Computational Linguistics.

\bibitem[{Thoppilan et~al.(2022)Thoppilan, De~Freitas, Hall, Shazeer,
  Kulshreshtha, Cheng, Jin, Bos, Baker, Du et~al.}]{thoppilan2022lamda}
Romal Thoppilan, Daniel De~Freitas, Jamie Hall, Noam Shazeer, Apoorv
  Kulshreshtha, Heng-Tze Cheng, Alicia Jin, Taylor Bos, Leslie Baker, Yu~Du,
  et~al. 2022.
\newblock Lamda: Language models for dialog applications.
\newblock \emph{arXiv preprint arXiv:2201.08239}.

\bibitem[{Touvron et~al.(2023{\natexlab{a}})Touvron, Lavril, Izacard, Martinet,
  Lachaux, Lacroix, Rozi{\`e}re, Goyal, Hambro, Azhar
  et~al.}]{touvron2023llama}
Hugo Touvron, Thibaut Lavril, Gautier Izacard, Xavier Martinet, Marie-Anne
  Lachaux, Timoth{\'e}e Lacroix, Baptiste Rozi{\`e}re, Naman Goyal, Eric
  Hambro, Faisal Azhar, et~al. 2023{\natexlab{a}}.
\newblock Llama: Open and efficient foundation language models.
\newblock \emph{arXiv preprint arXiv:2302.13971}.

\bibitem[{Touvron et~al.(2023{\natexlab{b}})Touvron, Martin, Stone, Albert,
  Almahairi, Babaei, Bashlykov, Batra, Bhargava, Bhosale
  et~al.}]{touvron2023llama2}
Hugo Touvron, Louis Martin, Kevin Stone, Peter Albert, Amjad Almahairi, Yasmine
  Babaei, Nikolay Bashlykov, Soumya Batra, Prajjwal Bhargava, Shruti Bhosale,
  et~al. 2023{\natexlab{b}}.
\newblock Llama 2: Open foundation and fine-tuned chat models.
\newblock \emph{arXiv preprint arXiv:2307.09288}.

\bibitem[{Wang et~al.(2023)Wang, Li, Chen, Zhu, Lin, Cao, Liu, Liu, and
  Sui}]{wang2023large}
Peiyi Wang, Lei Li, Liang Chen, Dawei Zhu, Binghuai Lin, Yunbo Cao, Qi~Liu,
  Tianyu Liu, and Zhifang Sui. 2023.
\newblock Large language models are not fair evaluators.
\newblock \emph{arXiv preprint arXiv:2305.17926}.

\bibitem[{West et~al.(2024)West, Lu, Dziri, Brahman, Li, Hwang, Jiang, Fisher,
  Ravichander, Chandu et~al.}]{west2023generative}
Peter West, Ximing Lu, Nouha Dziri, Faeze Brahman, Linjie Li, Jena~D Hwang,
  Liwei Jiang, Jillian Fisher, Abhilasha Ravichander, Khyathi Chandu, et~al.
  2024.
\newblock The generative ai paradox:“what it can create, it may not
  understand”.
\newblock In \emph{The Twelfth International Conference on Learning
  Representations}.

\bibitem[{Zellers et~al.(2018)Zellers, Bisk, Schwartz, and Choi}]{swag}
Rowan Zellers, Yonatan Bisk, Roy Schwartz, and Yejin Choi. 2018.
\newblock {SWAG}: A large-scale adversarial dataset for grounded commonsense
  inference.
\newblock In \emph{Proceedings of the 2018 Conference on Empirical Methods in
  Natural Language Processing}, pages 93--104, Brussels, Belgium. Association
  for Computational Linguistics.

\bibitem[{Zellers et~al.(2019)Zellers, Holtzman, Bisk, Farhadi, and Choi}]{hs}
Rowan Zellers, Ari Holtzman, Yonatan Bisk, Ali Farhadi, and Yejin Choi. 2019.
\newblock {H}ella{S}wag: Can a machine really finish your sentence?
\newblock In \emph{Proceedings of the 57th Annual Meeting of the Association
  for Computational Linguistics}, pages 4791--4800. Association for
  Computational Linguistics.

\bibitem[{Zheng et~al.(2023)Zheng, Zhou, Meng, Zhou, and
  Huang}]{zheng2023large}
Chujie Zheng, Hao Zhou, Fandong Meng, Jie Zhou, and Minlie Huang. 2023.
\newblock On large language models' selection bias in multi-choice questions.
\newblock \emph{arXiv preprint arXiv:2309.03882}.

\end{thebibliography}
